\documentclass{article}
\usepackage{ijcai24}

\usepackage{times}
\usepackage{soul}
\usepackage{url}
\usepackage[hidelinks]{hyperref}
\usepackage[utf8]{inputenc}
\usepackage[small]{caption}
\usepackage{graphicx}
\usepackage{amsmath}
\usepackage{amsthm}
\usepackage{booktabs}
\usepackage[switch]{lineno}

\usepackage[ruled,vlined]{algorithm2e}

\usepackage{color}
\usepackage{subfigure}
\usepackage{multirow}

\usepackage{xspace}

\makeatletter
\DeclareRobustCommand\onedot{\futurelet\@let@token\@onedot}
\def\@onedot{\ifx\@let@token.\else.\null\fi\xspace}

\def\ie{\emph{i.e}\onedot}

\makeatother
\usepackage{xcolor}

\title{DiverTEAM: An Efficient Evolutionary 
Algorithm\\for Diversified Top-$k$ (Weight) Clique Search Problems}

\author{
    Jiongzhi Zheng$^1$\thanks{The first two authors contribute equally.}\and 
    Jinghui Xue$^{1*}$\and 
    Kun He$^1$\thanks{Corresponding author. Email: brooklet60@hust.edu.cn.}\and 
    Chu-Min Li$^2$\and 
    Yanli Liu$^3$  
    \affiliations
    $^1$School of Computer Science and Technology, Huazhong University of Science and Technology, China\\
    $^2$MIS, University of Picardie Jules Verne, France\\
    $^3$WuHan University of Science and Technology, China
}

\begin{document}

\maketitle

\begin{abstract}
In many real-world problems and applications, finding only a single element, even though the best, among all possible candidates, cannot fully meet the requirements. We may wish to have a collection where each individual is not only outstanding but also distinctive. Diversified Top-$k$ (DT$k$) problems are a kind of combinatorial optimization problem for finding such a promising collection of multiple sub-structures, such as subgraphs like cliques and social communities. In this paper, we address two representative and practical DT$k$ problems, DT$k$ Clique search (DT$k$C) and DT$k$ Weight Clique search (DT$k$WC), and propose an efficient algorithm called \textbf{Diver}sified \textbf{T}op-$k$ \textbf{E}volutionary \textbf{A}lgorith\textbf{M} (\textbf{DiverTEAM}) for these two problems. DiverTEAM consists of a local search algorithm, which focuses on generating high-quality and \textit{diverse} individuals and sub-structures, and a genetic algorithm that makes individuals work as a \textit{team} and converge to (near-)optima efficiently. Extensive experiments show that DiverTEAM exhibits an excellent and robust performance across various benchmarks of DT$k$C and DT$k$WC.
\end{abstract}

\section{Introduction}
As a typical category of combinatorial optimization problems, the diversified Top-$k$ (DT$k$) problems aim to find (at most) $k$ diverse sub-structures whose combination maximizes (or minimizes) an objective function. For instance, many DT$k$ problems are defined on graphs, aiming to find (at most) $k$ subgraphs satisfying required demands and covering as many nodes or edges as possible, such as the DT$k$ (bi)clique search~\cite{yuan2016diversified,wu2020local,LQL+22}, DT$k$ community search~\cite{SWW+22}, DT$k$ subgraph querying~\cite{YFL16}, and DT$k$ edge patterns~\cite{edge-pattern} problems. Usually, the overlapping elements (such as nodes) are either ignored, penalized, or even prohibited in order to prioritize the diversity of solutions in DT$k$ problems. These DT$k$ problems have various real-world applications, such as pattern matching~\cite{WZ18}, route planning~\cite{LLZ+22}, best region searching~\cite{SOP+20}, etc. 

Among various DT$k$ problems defined on graphs, the DT$k$ Clique search (DT$k$C) and DT$k$ Weight Clique search (DT$k$WC) are two typical problems due to the representativeness of the clique model in various subgraph structures. Given an undirected graph, DT$k$C aims to find a collection of at most $k$ cliques covering as many vertices as possible. Given an undirected graph with each vertex associated with a positive weight, DT$k$WC aims to find a collection of at most $k$ cliques maximizing the total weight of covered vertices. Algorithms for DT$k$C and DT$k$WC can be categorized into  exact~\cite{zhou2021solving}, approximation~\cite{yuan2016diversified}, and heuristic algorithms, among which, heuristics appear to be more practical and efficient.

Wu et al.~\cite{wu2020local} proposed the first local search heuristic for DT$k$C, which updates the collection of cliques by constructing a new clique and then using it to replace a clique in the collection. Later on, similar local search heuristic methods are adopted to solve DT$k$WC~\cite{wu2021restart}. 
Recently, a hybrid evolutionary algorithm combining population-based and local search methods called HEA-D~\cite{wu2022head} was proposed for DT$k$WC, which significantly outperforms the existing local search and exact algorithms. HEA-D shows the potential of evolutionary algorithms in solving DT$k$WC. However, some deficient designs might limit the search capability of the evolutionary algorithm. For instance, the crossover and offspring selection approaches in HEA-D may cause the rapid loss of genes (\ie, cliques) and the population difficult to converge, and the local search might even degrade the population and further delay its convergence. Moreover, we found that no study has attempted to propose a heuristic algorithm to simultaneously solve the closely related DT$k$C and DT$k$WC problems, and the state-of-the-art DT$k$WC heuristic HEA-D shows 
low performance for DT$k$C.





To address the above issues and fill the above gap, we propose a novel and efficient hybrid evolutionary algorithm for both DT$k$C and DT$k$WC called \textbf{DiverTEAM} (\textbf{Diver}sified \textbf{T}op-$k$ \textbf{E}volutionary \textbf{A}lgorith\textbf{M}). Different from alternating local search and genetic algorithms in HEA-D, DiverTEAM separates them into two stages to make each more focused. 
In the first stage, DiverTEAM uses local search to focus on generating high-quality and diverse genes (\ie, cliques) and individuals (\ie, collections of cliques), during which the solutions are prohibited from getting worse. 
In the second stage, it uses a genetic algorithm to efficiently make the population converge to (near-)optimal solutions with our designed crossover operator, which can efficiently spread genes over the population and assign each individual suitable genes, making the individuals work as a team. 
Moreover, DiverTEAM does not fix the number of individuals generated by local search but allows it to adjust adaptively according to the scale and the $k$ value of the instances. 
In summary, the local search focuses on maximizing each team member's own contribution, while the genetic algorithm focuses on efficient and effective communication among the team members to maximize the team's contribution.

We further design some detailed methods to improve the algorithm's efficiency and performance, including a pseudo graph reduction preprocessing and several postprocessing. A solution-based tabu search is also incorporated to prevent duplicate searching (\ie, generating individuals that have occurred in history) and the population from converging slowly.

Existing studies for DT$k$WC and DT$k$C usually only evaluate algorithms on massive sparse Real-world graph benchmarks. To make a more convincing evaluation, we further consider the 2nd DIMACS graph benchmark that contains many dense graphs, as well as two kinds of random graphs, \ie, Erdős-Rényi (ER) graphs~\cite{erdHos1960evolution} and Barabási-Albert (BA) graphs~\cite{albert2002statistical}. 
Extensive experiments show that DiverTEAM performs 
excellently on various benchmarks for both DT$k$C and DT$k$WC, indicating the 
superiority of our algorithm. 

The main contributions of this work are as follows.

\begin{itemize}
\item We propose an efficient hybrid evolutionary algorithm 
called DiverTEAM for DT$k$C and DT$k$WC, consisting of a focused and adaptive local search algorithm and an effective genetic algorithm with efficient crossover operators. To our knowledge, this is the first heuristic algorithm proposed for both DT$k$C and DT$k$WC.
\item We propose several effective approaches, including the combination of solution-based tabu search methods and the genetic algorithm, a pseudo graph reduction preprocessing, and several postprocessing methods. These approaches, coupled with the proposed crossover operator and the evolutionary algorithm framework could also be used for other DT$k$ problems defined on graphs.
\item We evaluate algorithms on various datasets, including dense, massive sparse, and random graphs. Extensive experiments show that DiverTEAM significantly outperforms the state-of-the-art heuristics in various datasets of both DT$k$C and DT$k$WC, indicating its excellent performance and robustness. 
\end{itemize}

\section{Preliminaries}
\label{sec-Pre}
 
This section presents definitions of DT$k$C and DT$k$WC and provides an illustrative example for better understanding them.

Given an undirected graph $G = (V, E)$, where $V$ is the set of vertices and $E$ the set of edges, the density of $G$ is $2|E| / (|V|(|V|-1))$. Given a vertex set $V' \subseteq V$, $G[V']$ is defined as the subgraph induced by $V'$. For any vertex $v$ in $V$, we denote $N(v)$ as the set of vertices adjacent to $v$ in $G$. The degree of $v$ is $|N(v)|$. 
A clique $c$ in $G$ is a subset of $V$ such that for any two distinct vertices $u, v \in c$, edge $(u, v) \in E$. 
A clique set $C$ is a set of cliques, and the set of vertices covered by $C$ is defined as $cov(C) = \cup_{c \in C}{\left( \cup_{v \in c}\{v\} \right)}$. DT$k$C aims to find a clique set $C$ that contains at most $k$ cliques in $G$ and maximizes the number of covered vertices, \ie, $|cov(C)|$.

In DT$k$WC, each vertex $v \in V$ is assigned a positive weight, denoted as $w(v)$, and the weight of a clique $c$ is defined as $w(c) = \sum_{v \in c}w(v)$, \ie, the total weight of vertices in $c$, and the weight of a clique set $C$ is defined as $W(C) = \sum_{v \in cov(C)}w(v)$, \ie, the total weight of vertices covered by $C$. Given the above definitions, DT$k$WC is to find a clique set $C$ that contains at most $k$ cliques in $G$ with the maximum weight. DT$k$WC could be thought of as the weighted version of DT$k$C, and we regard $W(C)$ as their unified optimization objective for convenience by assigning a unit weight to each vertex in DT$k$C. 


Given a clique set $C$ and a clique $c \in C$, we define $priv(c, C) = c \backslash cov(C \backslash c)$ as the set of vertices in $c$ that are not covered by $C \backslash c$ and define $score(c, C) = \sum_{v \in priv(c, C)}w(v)$ as the reduction of $W(C)$ caused by removing $c$ from $C$.

\begin{figure}[!t]
    \centering
\includegraphics[width=0.8\columnwidth]{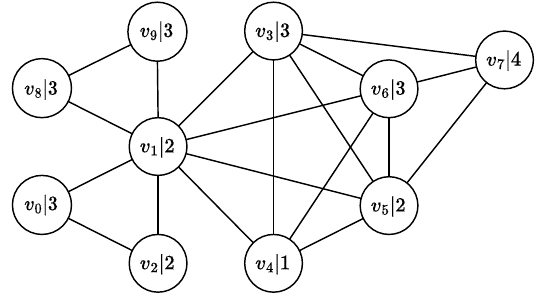}
    \caption{An instance for the DT$k$WC problem.\vspace{-0.5em}}
    \label{fig-instance}
\end{figure}

Figure~\ref{fig-instance} illustrates a graph $G = (V,E)$ for the DT$k$WC problem, where vertex $v_i|w$ represents vertex $v_i \in V$ with weight $w$. $G$ contains a number of high-quality cliques, such as $c_1 = \{v_1,v_3,v_4,v_5,v_6\}$, $c_2=\{v_1,v_8,v_9\}$, $c_3=\{v_0,v_1,v_2\}$, and $c_4=\{v_3,v_5,v_6,v_7\}$. When addressing the instance with $k=3$, the optimal clique set in $G$ is $C=\{c_1,c_2,c_3\}$ with set of its covered vertices $cov(C)= V\backslash \{v_7\}$ and its weight $W(C)=22$. Moreover, 
we have $priv(c_2,C)= \{v_0,v_2\}$ and $score(c_2,C)=5$.

\section{The DiverTEAM Evolutionary Algorithm}
This section introduces our DiverTEAM evolutionary algorithm proposed for DT$k$C and DT$k$WC. We first introduce the main process of DiverTEAM, and then present its components, including the preprocessing based on a pseudo graph reduction, the local search process, the genetic algorithm, and the postprocessing. 

\subsection{Main Process of DiverTEAM}
The main procedure of DiverTEAM is shown in Algorithm~\ref{alg_DiverTEAM}. The algorithm mainly consists of the separated local search and genetic algorithm stages. Before the first stage, the algorithm performs a pseudo graph reduction preprocessing (lines 2-4) to temporally reduce the isolated and leaf vertices (vertices with degree 0 or 1) from the input graph $G$. Then, the first stage (lines 5-8) uses a local search algorithm, \ie, function LocalSearch(), to generate individuals in the population based on the reduced graph $G'$. Once an individual is generated, a PostReduction() function will be called to try to improve the solution by considering the reduced vertices. In the second stage (lines 9-10), a genetic algorithm, \ie, function GeneticAlg() is used to search with the population. Finally, several postprocessing methods are applied in the PostProcessing() function (line 11) aming to further improve the individuals in the population.

The local search and genetic algorithm stages have their own stopping conditions (\textit{\uppercase\expandafter{\romannumeral1}} and \textit{\uppercase\expandafter{\romannumeral2}}). Suppose the cut-off time of the entire algorithm is $t_{max}$. In our implementation, we reserve 6 seconds for postprocessing to try to improve the most promising individuals. Thus the genetic algorithm stops when the running time reaches $t_{max} - 6$ (\ie, stopping condition \textit{\uppercase\expandafter{\romannumeral2}}). The local search process stops when the running time reaches $t_{max} - 16 - |P| \times k/10$ (\ie, stopping condition \textit{\uppercase\expandafter{\romannumeral1}}), where $|P|$ is the population size. In other words, we reverse $10 + |P| \times k/10$ seconds for the genetic algorithm because it needs more time to converge with larger $|P|$ and $k$. By associating the time limit of the genetic algorithm with $|P|$ and $k$, the number of individuals generated by the local search algorithm can be adjusted adaptively according to the scale and the $k$ value of the instances, making the algorithm more robust and effective.

\begin{algorithm}[t]
\fontsize{10.1pt}{15}
\caption{DiverTEAM}
\label{alg_DiverTEAM}
\LinesNumbered 
\KwIn{A graph $G$, an integer $k$, the maximum unimproved step in local search $M_{step}$, a time limit $t_{max}$}
\KwOut{A solution $C$}
Initialize the population $P \leftarrow \emptyset$\;
$IV \leftarrow$ the set of isolated vertices in $G$ with degree 0\;
$LV \leftarrow$ the set of leaf vertices in $G$ with degree 1\;
$G' \leftarrow G[V \backslash (IV \cup LV)]$\;
\While{stopping condition \uppercase\expandafter{\romannumeral1} is not met}{
$C \leftarrow$ LocalSearch($G'$, $k$, $M_{step}$)\;
$C \leftarrow$ PostReduction$(G,IV,LV,C)$\;
$P \leftarrow P \cup \{C\}$\;
}
\While{stopping condition \uppercase\expandafter{\romannumeral2} is not met}{
$P \leftarrow$ GeneticAlg($G$, $P$, $k$)\;
}
$P \leftarrow$ PostProcessing($G$, $P$, $t_{max}$)\;
$C^* \leftarrow \arg\max_{C \in P}{W(C)}$\;
\textbf{return} $C^*$\;
\end{algorithm}

\begin{algorithm}[t]
\fontsize{10.1pt}{15}
\caption{PostReduction($G,IV,LV,C$)}
\label{alg_PostReduction}
\LinesNumbered 
\KwIn{A graph $G$, the set of isolated vertices $IV$, the set of leaf vertices $LV$, a clique set $C$}
\KwOut{A clique set $C$}
\For{\rm{\textbf{each}} vertex $v \in (IV \cup LV)$}{
$c' \leftarrow N(v) \cup \{v\}$\;
\If{$w(c') > \min\{\cup_{c \in C}{score(c,C)}$\}
}{$C \leftarrow C \cup \{c'\}$\;
$c_{min} \leftarrow \arg\min_{c \in C}{score(c, C)}$\;
$C \leftarrow C \backslash c_{min}$\;}
}
\textbf{return} $C$\;
\end{algorithm}

\subsection{Pseudo Graph Reduction Preprocessing}
In many real-world applications, such as social networks, graphs are massive and sparse. Thus, searching the entire graph is inefficient, and many graph reduction methods have been proposed for various problems over graphs~\cite{ZHX+21,XLY22}. Due to the complexity of DT$k$C and DT$k$WC, it is very hard to reduce the graph with theoretical guarantees. Inspired by the leaf vertex union match method~\cite{Zhou0ZL022} that ignores leaf vertices (with degree 1) and handles them when encountering their parent vertices, we propose a pseudo graph reduction method that ignores isolated vertices (with degree 0) and leaf vertices during each local search process and considers them after generating each individual, so as to improve the search efficiency.

We denote the set of all isolated vertices as $IV$ and the set of all leaf vertices as $LV$. During the LocalSearch() function (line 6 in Algorithm~\ref{alg_DiverTEAM}), vertices in $IV$ and $LV$ are not considered. Once the LocalSearch() function returns a solution $C$, we utilize a PostReduction() function to enhance the quality of $C$ by taking into account the ignored vertices, which is depicted in Algorithm~\ref{alg_PostReduction}. It iterates through each isolated or leaf vertex and identifies the corresponding clique $c'$ with size 1 or 2 (lines 1-2), 
trying to replace one of the cliques in $C$ with $c'$ for improvement 
(lines 3-6).

\begin{algorithm}[t]
\fontsize{10.1pt}{15}
\caption{LocalSearch($G$, $k$, $M_{step}$)}
\label{alg_LocalSearch}
\LinesNumbered 
\KwIn{A graph $G$, an integer $k$, the maximum unimproved step in local search $M_{step}$}
\KwOut{A solution $C$}
Initialize $C \leftarrow \emptyset$\;
\For{$i \leftarrow 1 : k$}{
$c \leftarrow$ FindClique($G$)\;
$C \leftarrow C \cup \{c\}$\;
}
Initialize $step \leftarrow 0$\;
\While{$step < M_{step}$}{
$step \leftarrow step + 1$\;
$c \leftarrow$ FindClique($G$), $C' \leftarrow C \cup \{c\}$\;
$c_{min} \leftarrow \arg\min_{c \in C'}{score(c, C')}$\;
$C' \leftarrow C' \backslash c_{min}$\;
\If{$W(C') > W(C)$}{$C \leftarrow C'$\;
$step \leftarrow 0$\;}

}
\textbf{return} $C$\;
\end{algorithm}

\subsection{Local Search Process}
The procedure of the local search algorithm in DiverTEAM is presented in Algorithm~\ref{alg_LocalSearch}. The algorithm first constructs a solution $C$ by merging $k$ cliques (lines 1-4). Each clique is found by function FindClique() (line 3), which first samples a starting vertex as the initial clique and then extends it as HEA-D~\cite{wu2022head} does. Then, the algorithm iteratively adds a clique found by FindClique() into $C$ and removes the clique with the minimum $score$ from $C$ until the maximum unimproved step $M_{step}$ is reached (line 6).

Note that during the local search process, solution $C$ never becomes worse because the added and removed cliques can be the same. Since DT$k$C and DT$k$WC usually have huge solution spaces, different from HEA-D using simulated annealing methods to accept worse solutions, we prohibit the solution from getting worse during the local search, which can prevent the algorithm from being hard to converge. Since the genetic algorithm in DiverTEAM is very efficient, most of the running time (usually more than 3/4 of the total running time) is spent on local search in DiverTEAM, making the local search process concentrate on generating high-quality and diverse local optimal solutions for the population.


\begin{algorithm}[t]
\fontsize{10.1pt}{15}
\caption{GeneticAlg($G$, $P$, $k$)}
\label{alg_GeneticAlg}
\LinesNumbered 
\KwIn{A graph $G$, a population $P$, an integer $k$}
\KwOut{A population $P$}

Randomly shuffle $P$\;



\For{$i \leftarrow 1 : |P|$}{

$C_1 \leftarrow i$-th individual in $P$\;

$C_2 \leftarrow (i \mod |P|+1)$-th individual in $P$\;

Initialize $score^* \leftarrow -\infty$, $c_1^* \leftarrow \emptyset$, $c_2^* \leftarrow \emptyset$\;

\For{\rm{\textbf{each}} clique $c_1 \in C_1$}{
\For{\rm{\textbf{each}} clique $c_2 \in C_2$}{

$score \leftarrow W(C_1\backslash \{c_1\} \cup \{c_2\}) - W(C_1)$\;

\If{$score > score^*$}{
    $score^* \leftarrow score$, $c_1^* \leftarrow c_1$, $c_2^* \leftarrow c_2$\;
}
}
}

$C_1 \leftarrow C_1\backslash \{c_1^*\} \cup \{c_2^*\}$\;

}

\textbf{return} $P$\;
\end{algorithm}

\begin{algorithm}[t]
\fontsize{10.1pt}{15}
\caption{PostProcessing($G,P,t_{max}$)}
\label{alg_postprocessing}
\LinesNumbered 
\KwIn{A graph $G=(V,E)$, a population $P$, a time limit $t_{max}$}
\KwOut{A population $P$}
Sort $P$ in descending order of the individuals' weights\;
$i \leftarrow 1$\;
\While{$t_{max}$ is not reached}{
$C \leftarrow$ the $i$-th individual in $P$\;
$C' \leftarrow \emptyset$\;
\For{\rm{\textbf{each}} clique $c\in C$ }{
    $c \leftarrow c \backslash cov(C')$\;
    Expand $c$ to maximal, prioritizing vertices that are not in $cov(C')$\;
    $C' \leftarrow C'\cup \{c\}$\;
}
\For{\rm{\textbf{each}} vertex $v \in V \backslash cov(C')$}{
\For{\rm{\textbf{each}} clique $c\in C'$ }{
$c' \leftarrow c \backslash \left(V \backslash N(v)\right) \cup \{v\} $\;
\If{$w(C'\backslash\{c\}\cup \{c'\}) > w(C')$}{
    $C'\leftarrow C'\backslash\{c\}\cup\{c'\}$\;
    \textbf{break}\;
}
}
}
Replace $C$ in $P$ with $C'$\;
$i \leftarrow i+1$\;
}
\textbf{return} $P$\;
\end{algorithm}

\subsection{Genetic Algorithm}

The procedure of the genetic algorithm in DiverTEAM is presented in Algorithm~\ref{alg_GeneticAlg}.
At each generation, each individual is selected once as parent $C_1$ and once as parent $C_2$ in a random order (lines 1-4). For each pair of parents, $C_1$ and $C_2$, an efficient crossover operator is proposed for merging their genes and generating high-quality offspring. In DiverTEAM, we regard the cliques in each individual as genes. The crossover operator actually tries to replace a clique $c_1 \in C_1$ with a clique $c_2 \in C_2$ to evolve $C_1$, and it traverses all possible replacements to find the best one (lines 5-11).


Note that we have found sufficient high-quality and diverse genes (i.e., cliques), which are distributed among various individuals in the population, through the local search process.
The goal of the genetic algorithm is to effectively propagate these promising genes throughout the entire population, enabling each individual to discover suitable genes that match their characteristics efficiently. 
Importantly, the time complexity of the GeneticAlg() function is $O(|P|k^2)$, allowing us to propagate the genes and evolve the population efficiently. Experimental results also demonstrate the superiority of our proposed genetic algorithm over that in HEA-D.



\subsection{Tabu Search in the Genetic Algorithm}
Solution-based tabu search methods are widely used in heuristics (mainly local search) for various combinatorial optimization problems, such as the inventory routing problem~\cite{SuTabu} and $p$-next center problem~\cite{ZSL+22}. In DiverTEAM, we apply solution-based tabu search to the genetic algorithm to prevent the algorithms from searching for the solutions that have been previously encountered and ensure the diversity of the population.

Specifically, we establish three hash vectors $H_1, H_2, H_3$ to represent the tabu list. Each vector has a length of $L=10^8$. The three hash vectors are initialized to 0. Additionally, we define three hash functions, denoted as $h_1(\cdot), h_2(\cdot), h_3(\cdot)$, each maps a solution $C$ within the range $[0, L)$. A solution $C$ is considered in the tabu list if $\wedge_{i=1}^{3}H_i(h_i(C)) = 1$. We associate three random integers $j_1(v), j_2(v), j_3(v)$ with each vertex $v$, the hash value $h_i(C), i \in \{1,2,3\}$, can be calculated by $h_i(C) = \sum_{c\in C}\sum_{v\in c} j_i(v) \mod L$.

In the genetic algorithm, each solution $C$ that occurred in the population will be added to the tabu list by setting $\wedge_{i=1}^{3}H_i(h_i(C)) = 1$, and any solution $C$ prohibited by the tabu list will not occur in future populations.

\subsection{Postprocessing}

We propose two postprocessing methods aiming to further improve the individuals in the population after the genetic algorithm, as depicted in Algorithm~\ref{alg_postprocessing}. The algorithm tries to improve the most promising individuals until the time limit is reached. The first postprocessing method is a construction heuristic (lines 6-9), which starts from an empty set and iteratively constructs a maximal clique based on parts of each clique $c$ in the individual $C$. For each clique $c \in C$, the algorithm first removes vertices covered by the current constructed solution $C'$ from $c$, and then expands it to a maximal clique, preferring vertices not covered by $C'$. The second method is a searching heuristic (lines 10-15), which actually tries to add uncovered vertices to the solution for possible improvement.

Note that neither of these two postprocessing methods makes the solution worse. However, they may degrade the quality of the genes and are time-consuming. Thus, we do not use them frequently during the entire algorithm. 



\section{Experimental Results}
\begin{table*}[!t]
\centering
\footnotesize
\resizebox{\linewidth}{!}{
\begin{tabular}{l|c|cccc|cccc|cccc} \toprule
\multirow{2}{*}{Comparison} &
  \multirow{2}{*}{$k$} &
  \multicolumn{4}{c|}{Real-world} &
  \multicolumn{4}{c|}{2nd DIMACS} &
  \multicolumn{4}{c}{Random graph} \\
&    & $N^+_{best}$         & $N^-_{best}$ & $N^+_{avg}$         & $N^-_{avg}$  & $N^+_{best}$         & $N^-_{best}$ & $N^+_{avg}$       & $N^-_{avg}$ & $N^+_{best}$         & $N^-_{best}$ & $N^+_{avg}$         & $N^-_{avg}$ \\ \hline
\multirow{5}{*}{DiverTEAM vs. HEA-D} &
  10 &
  \textbf{41} &
  1 &
  \textbf{45} &
  5 &
  \textbf{50} &
  1 &
  \textbf{50} &
  4 &
  \textbf{37} &
  4 &
  \textbf{43} &
  5 \\
 & 20 & \textbf{46} & 4    & \textbf{52} & 8    & \textbf{45} & 1           & \textbf{45} & 1    & \textbf{44} & 3    & \textbf{47} & 3    \\
 & 30 & \textbf{51} & 1    & \textbf{57} & 4    & \textbf{33} & 1           & \textbf{35} & 3    & \textbf{47} & 2    & \textbf{48} & 4    \\
 & 40 & \textbf{54} & 2    & \textbf{63} & 4    & \textbf{27} & 1          & \textbf{28} & 2    & \textbf{48} & 1    & \textbf{50} & 2    \\
 & 50 & \textbf{58} & 2    & \textbf{66} & 4    & \textbf{25} & 1           & \textbf{24} & 3    & \textbf{48} & 1    & \textbf{50} & 2    \\ \hline
\multirow{5}{*}{DiverTEAM vs. TOPKLS} &
  10 &
  \textbf{23} &
  6 &
  \textbf{23} &
  12 &
  \textbf{33} &
  23 &
  \textbf{39} &
  27 &
  \textbf{24} &
  14 &
  \textbf{25} &
  16 \\
 & 20 & \textbf{35} & 3    & \textbf{37} & 10   & \textbf{33}          & 19 & \textbf{41} & 17   & \textbf{24} & 16   & \textbf{25} & 18   \\
 & 30 & \textbf{41} & 2    & \textbf{46} & 8    & \textbf{33} & 14          & \textbf{36} & 13   & \textbf{26} & 14   & \textbf{28} & 14   \\
 & 40 & \textbf{45} & 4    & \textbf{56} & 2    & \textbf{31} & 7          & \textbf{40} & 7   & \textbf{29} & 14   & \textbf{29} & 14   \\
 & 50 & \textbf{52} & 2    & \textbf{65} & 2    & \textbf{25} & 6          & \textbf{36} & 6   & \textbf{28} & 14   & \textbf{30} & 15  \\ \bottomrule
\end{tabular}} 
\caption{Comparison between DiverTEAM and two baselines, HEA-D and TOPKLS, on the DT$k$C problem.} 
\label{Table-compare-wo-weight}
\end{table*}

\begin{table*}[!t]
\centering
\footnotesize
\resizebox{\linewidth}{!}{
\begin{tabular}{l|c|cccc|cccc|cccc} \toprule
\multirow{2}{*}{Comparison} &
  \multirow{2}{*}{$k$} &
  \multicolumn{4}{c|}{Real-world} &
  \multicolumn{4}{c|}{2nd DIMACS} &
  \multicolumn{4}{c}{Random graph} \\
                              &    & $N^+_{best}$         & $N^-_{best}$ & $N^+_{avg}$         & $N^-_{avg}$  & $N^+_{best}$         & $N^-_{best}$ & $N^+_{avg}$       & $N^-_{avg}$ & $N^+_{best}$         & $N^-_{best}$ & $N^+_{avg}$         & $N^-_{avg}$ \\ \hline
\multirow{5}{*}{DiverTEAM vs. HEA-D}    & 10 & \textbf{49} & 2    & \textbf{53} & 4    & \textbf{52} & 2    & \textbf{54} & 3    & \textbf{39} & 0    & \textbf{43} & 0    \\
                              & 20 & \textbf{53} & 1    & \textbf{59} & 2    & \textbf{47} & 0    & \textbf{47} & 0    & \textbf{41} & 0    & \textbf{45} & 0    \\
                              & 30 & \textbf{55} & 1    & \textbf{63} & 2    & \textbf{36} & 0    & \textbf{36} & 3    & \textbf{41} & 0    & \textbf{48} & 0    \\
                              & 40 & \textbf{57} & 1    & \textbf{65} & 2    & \textbf{28} & 0    & \textbf{29} & 1    & \textbf{46} & 0    & \textbf{52} & 0    \\
                              & 50 & \textbf{59} & 1    & \textbf{71} & 2    & \textbf{25} & 1    & \textbf{26} & 2    & \textbf{44} & 0    & \textbf{54} & 0    \\ \hline
\multirow{5}{*}{DiverTEAM vs. TOPKWCLQ} & 10 & \textbf{72} & 0    & \textbf{84} & 0    & \textbf{63} & 0    & \textbf{64} & 0    & \textbf{44} & 2    & \textbf{52} & 2    \\
                              & 20 & \textbf{82} & 0    & \textbf{87} & 0    & \textbf{53} & 0    & \textbf{56} & 0    & \textbf{49} & 1    & \textbf{59} & 1    \\
                              & 30 & \textbf{85} & 0    & \textbf{91} & 0    & \textbf{46} & 0    & \textbf{48} & 0    & \textbf{52} & 3    & \textbf{61} & 2    \\
                              & 40 & \textbf{85} & 0    & \textbf{91} & 0    & \textbf{40} & 0    & \textbf{43} & 0    & \textbf{57} & 3    & \textbf{61} & 2    \\
                              & 50 & \textbf{85} & 0    & \textbf{89} & 0    & \textbf{33} & 0    & \textbf{34} & 0    & \textbf{59} & 2    & \textbf{62} & 2  \\ \bottomrule 
\end{tabular}}
\caption{Comparison between DiverTEAM and two baselines, HEA-D and TOPKWCLQ, on the DT$k$WC problem.}
\label{Table-compare-with-weight}
\end{table*}

Experimental results consist of two parts. We first evaluate the overall performance of the proposed DiverTEAM\footnote{The source codes of DiverTEAM are available at https://github.com/[MASKED-FOR-REVIEW].} algorithm on the DT$k$C and DT$k$WC problems. We select the state-of-the-art hybrid evolutionary algorithm, HEA-D~\cite{wu2022head}, and a local search algorithm, TOPKWCLQ~\cite{wu2021restart}, as the baselines for DT$k$WC, and HEA-D and a local search algorithm, TOPKLS~\cite{wu2020local}, as the baselines for DT$k$C. Then, we perform extensive ablation studies by comparing DiverTEAM with its various variants to evaluate the effectiveness of its components. Results of the baselines are obtained by running their source codes. 



\subsection{Experimental Setup}
All the algorithms were implemented in C++ and executed on a server with an AMD EPYC 7H12 CPU, running Ubuntu 18.04 Linux operating system. 
To make a convincing evaluation of DiverTEAM, we conducted experiments using three benchmark datasets, including the Real-world benchmark\footnote{http://lcs.ios.ac.cn/\%7Ecaisw/Resource/realworld\%20\\graphs.tar.gz} that contains 102 real-world sparse graphs sourced from the Network Data Repository~\cite{RA15} and widely used by the baselines, the 2nd DIMACS benchmark\footnote{http://archive.dimacs.rutgers.edu/pub/challenge/graph/\\benchmarks/clique/} that consists of 80 almost dense graphs with up to 4,000 vertices and densities ranging from 0.03 to 0.99, and a random benchmark comprising 30 Erdős-Rényi (ER) graphs~\cite{erdHos1960evolution} and 35 Barabási-Albert (BA) graphs~\cite{albert2002statistical}. The number of vertices of the 65 random graphs varies from 1,000, 2,000, 4,000, 8,000, and 16,000. For each number of vertices, we generated six ER graphs with densities of 0.001, 0.01, 0.05, 0.1, 0.2, and 0.4, and seven BA graphs with each new vertex adjacent to 1, 10, 50, 100, 200, 400, or 800 edges with the existing vertices.




For each graph, we generate 5 DT$k$C (DT$k$WC) instances with $k = [10, 20, 30, 40, 50]$. For each DT$k$WC instance, the $i$-th vertex is assigned with a weight $(i \mod 200) + 1$ as DT$k$WC baselines and many studies about weighted clique do~\cite{Weight1,JiangLLM18,Weight2}. Moreover, each algorithm is performed on 10 independent runs with a cut-off time of 600 seconds on each instance, as the baselines do.


\subsection{Performance Evaluation}

\begin{figure*}[!t]
\centering
\subfigure[]{
\includegraphics[width=0.49\columnwidth]{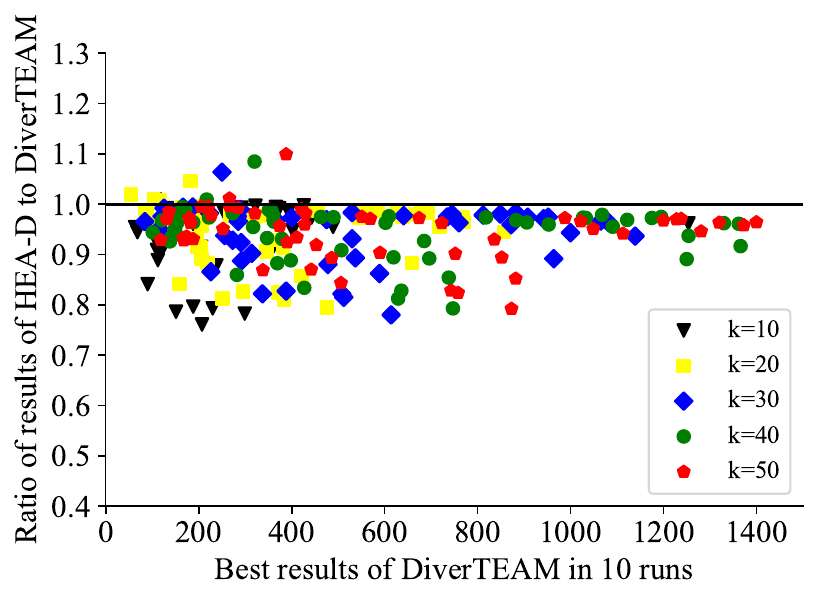}
\label{fig-HEAD-Ours-real-wo-weight}}
\subfigure[]{
\includegraphics[width=0.49\columnwidth]{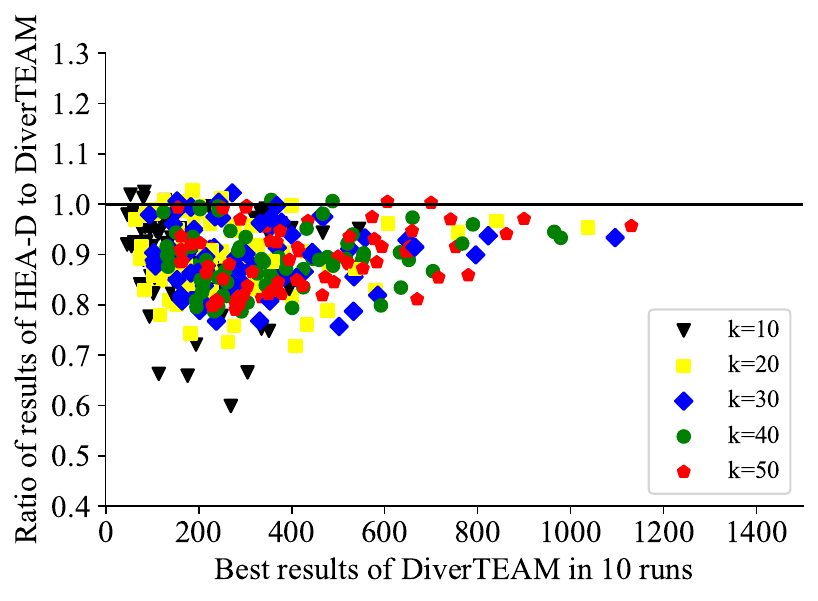}
\label{fig-HEAD-Ours-DIMACS2&random-wo-weight}}
\subfigure[]{
\includegraphics[width=0.49\columnwidth]{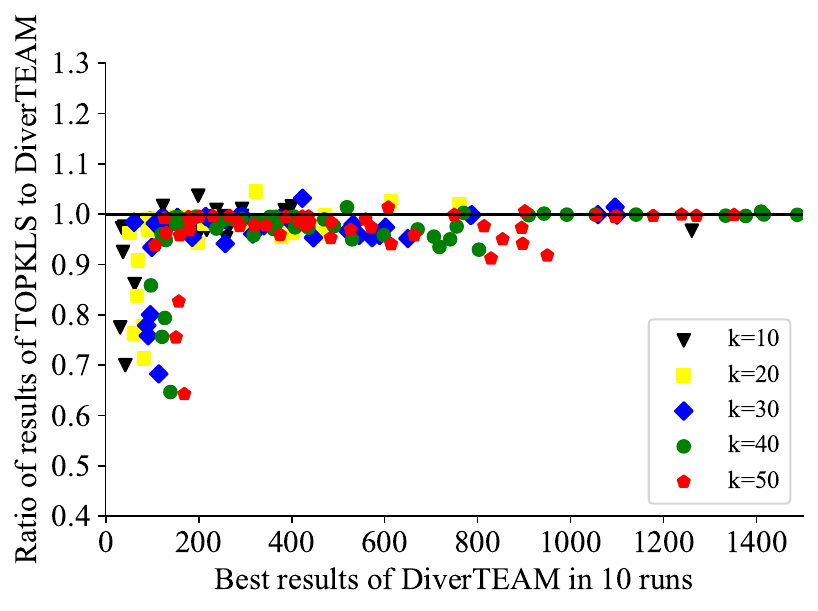}
\label{fig-Topkwclq-Ours-real-wo-weight}}
\subfigure[]{
\includegraphics[width=0.49\columnwidth]{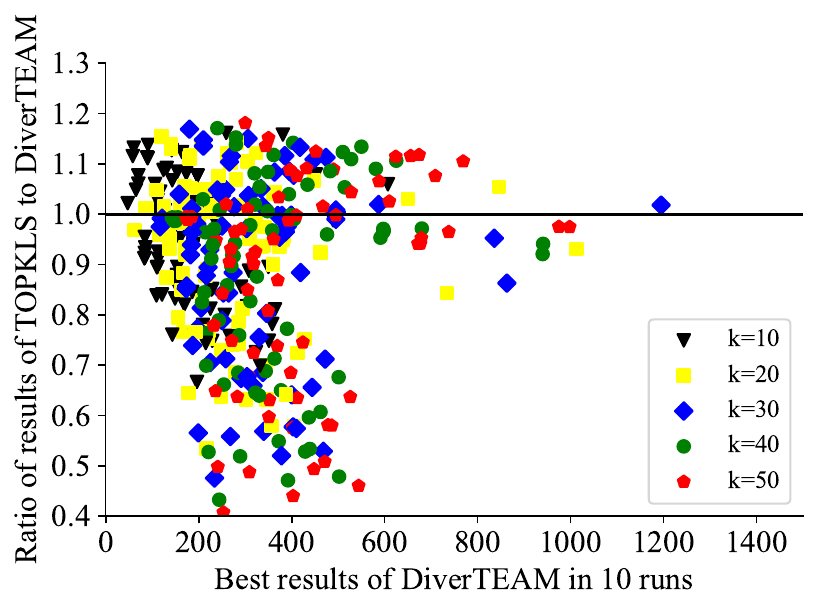}
\label{fig-Topkwclq-DIMACS2&random-wo-weight}}

\caption{Comparison of DiverTEAM and DT$k$C baselines. A point with coordinates $(x,y)$ 
represents a DT$k$C instance that the best result obtained by DiverTEAM is $x$ 
and the baseline algorithm is $x \times y$ in 10 runs. (a) DiverTEAM vs. HEA-D on Real-world graphs; (b) DiverTEAM vs. HEA-D on 2nd DIMACS and random graphs; (c) DiverTEAM vs. TOPKLS on Real-world graphs; (d) DiverTEAM vs. TOPKLS on 2nd DIMACS and random graphs.\vspace{-0.5em}}

\label{fig-scatter-wo-weight}
\end{figure*}

 \begin{figure*}[!t]
\centering
\subfigure[]{
\includegraphics[width=0.49\columnwidth]{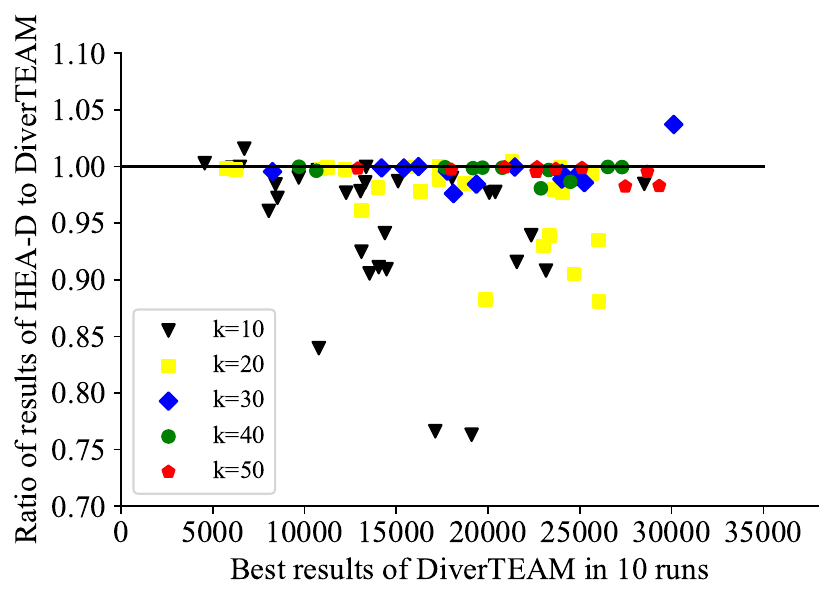}
\label{fig-HEAD-Ours-real}}
\subfigure[]{
\includegraphics[width=0.49\columnwidth]{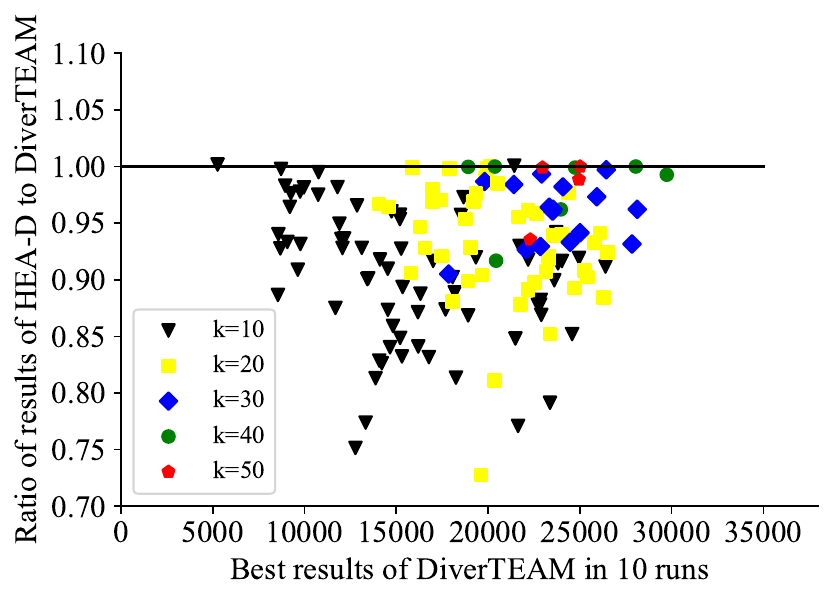}
\label{fig-HEAD-Ours-DIMACS2&random}}
\subfigure[]{
\includegraphics[width=0.49\columnwidth]{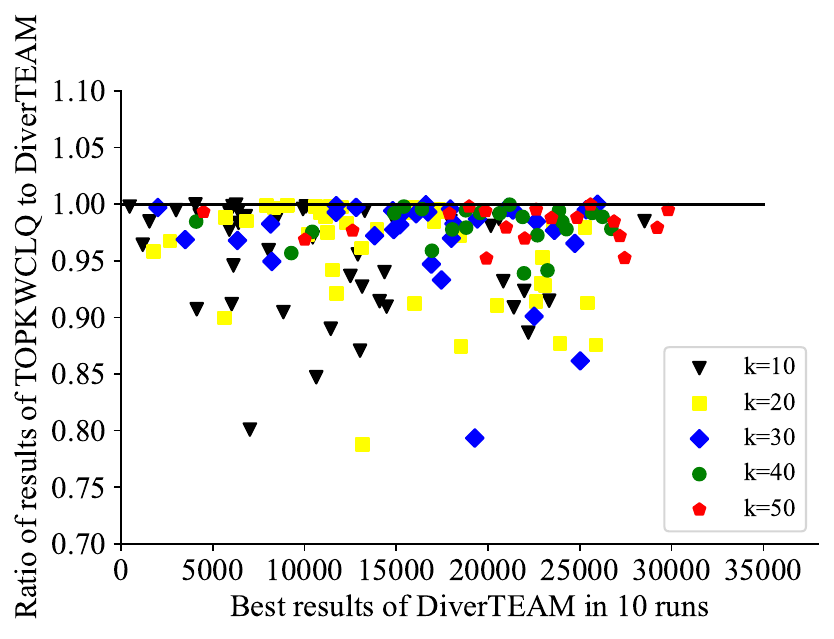}
\label{fig-Topkwclq-Ours-real}}
\subfigure[]{
\includegraphics[width=0.49\columnwidth]{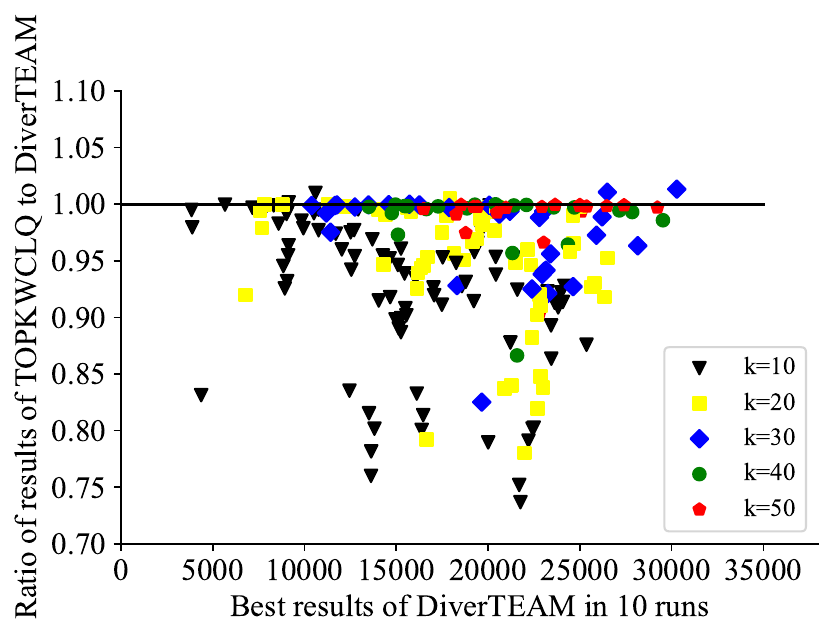}
\label{fig-Topkwclq-DIMACS2&random}}

\caption{Comparison of DiverTEAM and DT$k$WC baselines. A point with coordinates $(x,y)$ 
represents a DT$k$WC instance that the best result obtained by DiverTEAM is $x$ 
and the baseline algorithm is $x \times y$ in 10 runs. (a) DiverTEAM vs. HEA-D on Real-world graphs; (b) DiverTEAM vs. HEA-D on 2nd DIMACS and random graphs; (c) DiverTEAM vs. TOPKWCLQ on Real-world graphs; (d) DiverTEAM vs. TOPKWCLQ on 2nd DIMACS and random graphs.}
\label{fig-scatter-weight}
\end{figure*}

Tables \ref{Table-compare-wo-weight} and \ref{Table-compare-with-weight}
summarizes the comparison results between DiverTEAM and the baseline algorithms on DT$k$C and DT$k$WC instances based on the three benchmarks. Column $N^+_{best}$ (resp. $N^+_{avg}$) indicates the number of instances that DiverTEAM obtains better results than 
the compared algorithm in terms of the best (resp. average) solutions in 10 runs. Column $N^-_{best}$ (resp. $N^-_{avg}$) indicates the number of instances that DiverTEAM obtains worse results than the compared algorithm in terms of the best (resp. average) solutions in 10 runs. The best results in the tables appear in bold.

The results show that DiverTEAM significantly outperforms the baseline local search and hybrid evolutionary algorithms over all the DT$k$C and DT$k$WC benchmarks. TOPKLS also shows good performance on DT$k$C instances, especially on instances with small $k$ values. This is because DT$k$C instances based on small and dense 2nd DIMACS graphs with small $k$ values have relatively small solution spaces, which are more suitable for the local search algorithm than the evolutionary algorithm. The results also indicate that HEA-D is not very good at solving DT$k$C instances compared to TOPKLS, indicating that the closely related DT$k$C and DT$k$WC problems are distinct in suitable solving methods. With the benefit of our framework that separates local search and genetic algorithms to make each more focused, the effective genetic algorithm with efficient crossover operators, and techniques including solution-based tabu search, preprocessing, and postprocessing, our proposed DiverTEAM becomes a generic algorithm that has excellent performance and robustness on various benchmarks of the two problems. 
Followup ablation studies in Section~\ref{sec-ablation} also demonstrate the advantages of the above components and techniques of DiverTEAM.

\begin{table*}[!t]
\centering
\footnotesize
\resizebox{\linewidth}{!}{
\begin{tabular}{l|cccc|cccc|cccc} \toprule
\multicolumn{1}{l|}{\multirow{2}{*}{DiverTEAM vs.}} &
  \multicolumn{4}{c|}{Real-world} &
  \multicolumn{4}{c|}{2nd DIMACS} &
  \multicolumn{4}{c}{Random graph} \\
\multicolumn{1}{c|}{} &
  $N^+_{best}$         & $N^-_{best}$ & $N^+_{avg}$         & $N^-_{avg}$  & $N^+_{best}$         & $N^-_{best}$ & $N^+_{avg}$       & $N^-_{avg}$ & $N^+_{best}$         & $N^-_{best}$ & $N^+_{avg}$         & $N^-_{avg}$ \\ \hline
DiverTEAM$_\text{pre}^-$ &
  \textbf{10} &
  5 &
  12 &
  \textbf{17} &
  \textbf{15} &
  13 &
  \textbf{28}&
  9 &
  \textbf{20} &
  10 &
  \textbf{34} &
  9 \\
DiverTEAM$_\text{tabu}^-$   & \textbf{9}  & 3 & \textbf{23} & 8  & \textbf{18} & 3 & \textbf{26} & 10 & \textbf{18} & 9 & \textbf{30} & 8  \\
DiverTEAM$_\text{GA}^-$         & \textbf{28} & 2 & \textbf{43} & 0  & \textbf{18} & 9  & \textbf{26} & 11  & \textbf{29} & 2 & \textbf{36} & 3  \\
DiverTEAM$_\text{HEA-D}^+$  & \textbf{16} & 2 & \textbf{32} & 5  & \textbf{16} & 11 & \textbf{24} & 14 & \textbf{15} & 5 & \textbf{26} & 15 \\
DiverTEAM$_\text{post}^-$
& \textbf{11} & 2 & \textbf{18} & 11 & \textbf{31} & 3  & \textbf{38} & 1  & \textbf{29} & 3 & \textbf{37} & 3 \\ \bottomrule 
\end{tabular}}
\caption{Comparison of DiverTEAM and its five variants on DT$k$C problem based on three benchmarks with $k = 30$.}
\label{Table-abalation-wo-weight}
\end{table*}
\begin{table*}[!t]
\centering
\footnotesize
\resizebox{\linewidth}{!}{
\begin{tabular}{l|cccc|cccc|cccc} \toprule
\multicolumn{1}{l|}{\multirow{2}{*}{DiverTEAM vs.}} &
  \multicolumn{4}{c|}{Real-world} &
  \multicolumn{4}{c|}{2nd DIMACS} &
  \multicolumn{4}{c}{Random graph} \\
\multicolumn{1}{c|}{} & $N^+_{best}$         & $N^-_{best}$ & $N^+_{avg}$         & $N^-_{avg}$  & $N^+_{best}$         & $N^-_{best}$ & $N^+_{avg}$       & $N^-_{avg}$ & $N^+_{best}$         & $N^-_{best}$ & $N^+_{avg}$         & $N^-_{avg}$        \\ \hline
DiverTEAM$_\text{pre}^-$ &
  \textbf{22} &
  6 &
  \textbf{27} &
  17 &
  \textbf{17} &
  11 &
  \textbf{20} &
  17 &
  \textbf{20} &
  19 &
  \textbf{28} &
  14 \\
DiverTEAM$_\text{tabu}^-$     & \textbf{22} & 8    & \textbf{32} & 12   & \textbf{23} & 7    & \textbf{27} & 10   & \textbf{24} & 16   & \textbf{39} & 4           \\
DiverTEAM$_\text{GA}^-$           & \textbf{37} & 1    & \textbf{53} & 1    & \textbf{25} & 6    & \textbf{28} & 11   & \textbf{35} & 5    & \textbf{39} & 4           \\
DiverTEAM$_\text{HEA-D}^+$    & \textbf{29} & 5    & \textbf{49} & 3    & \textbf{24} & 7    & \textbf{22} & 17   & \textbf{21} & 17   & 21          & \textbf{22} \\
DiverTEAM$_\text{post}^-$      & \textbf{29} & 2    & \textbf{38} & 7    & \textbf{43} & 0    & \textbf{43} & 0    & \textbf{33} & 5    & \textbf{39} & 2 \\ \bottomrule           
\end{tabular}}
\caption{Comparison of DiverTEAM and its five variants on DT$k$WC problem based on three benchmarks with $k = 30$.}
\label{Table-abalation-with-weight}
\end{table*}

To clearly present the gap in results of DiverTEAM and the baseline algorithms, we further present scatter plots on DT$k$C and DT$k$WC instances in Figures \ref{fig-scatter-wo-weight} and \ref{fig-scatter-weight}, respectively, to depict the detailed results. 
We only present results in instances where DiverTEAM and the corresponding baseline obtain different best results in 10 runs and also abandon a few outliers for DT$k$C (resp. DT$k$WC) instances where the objective values of the best results found by the algorithms are larger than 2,000 (resp. 30,000).

The results in Figures~\ref{fig-scatter-wo-weight} and \ref{fig-scatter-weight} are consistent with the results in Tables~\ref{Table-compare-wo-weight} and \ref{Table-compare-with-weight}, and can further reveal that there exists a large gap between the results obtained by DiverTEAM and the baselines in many instances. The objective values of the best results obtained by the baselines are only $70\%$ of those by DiverTEAM for many instances, indicating a significant superiority of DiverTEAM over the baselines. Moreover, the baselines show unstable performance on different benchmarks. For example, the objective values of most results obtained by HEA-D in Real-world DT$k$WC instances surpass $90\%$ of those by DiverTEAM, while on many 2nd DIMACS and random graphs, HEA-D can only yield about $80\%$ of the objective values of results of DiverTEAM. The local search algorithm TOPKLS exhibits notable instability on 2nd DIMACS and random graphs, which can yield results $20\%$ times better than that of DiverTEAM sometimes and also can only yield results with half of the objective values 
as compared to DiverTEAM in many instances. The results indicate again that DiverTEAM has excellent performance and robustness in various instances of DT$k$C and DT$k$WC.


\subsection{Ablation Study}
\label{sec-ablation}

To evaluate the effectiveness of components and techniques in DiverTEAM, we compare DiverTEAM with its five variant algorithms on all the DT$k$C and DT$k$WC instances with $k = 30$. The variants include DiverTEAM$_\text{pre}^-$, which removes the pseudo graph reduction preprocessing, DiverTEAM$_\text{tabu}^-$, which removes the solution-based tabu search in the genetic algorithm, DiverTEAM$_\text{GA}^-$, which removes the genetic algorithm and uses the local search algorithm to calculate before reaching stopping condition \textit{II}, DiverTEAM$_\text{HEA-D}^+$, which replaces the genetic algorithm in DiverTEAM with that in HEA-D, and DiverTEAM$_\text{post}^-$, which removes the postprocessing methods. The results on the  
two problems are shown in Tables \ref{Table-abalation-wo-weight} and \ref{Table-abalation-with-weight}, respectively, with best results appearing in bold.


The results reveal that DiverTEAM yields considerably better performance than the five variants in general. DiverTEAM outperforms DiverTEAM$_\text{pre}^-$, indicating that the proposed pseudo graph reduction method can improve the performance by reducing the search space. DiverTEAM outperforms DiverTEAM$_\text{tabu}^-$, demonstrating that the solution-based tabu search can prevent the algorithm from searching for the same solutions and help the algorithm obtain better results. DiverTEAM performs better than DiverTEAM$_\text{GA}^-$, indicating that the genetic algorithm can significantly improve the solutions generated by the local search algorithm. DiverTEAM performs better than DiverTEAM$_\text{HEA-D}^+$, indicating the superiority of our proposed genetic algorithm over that in HEA-D. Moreover, DiverTEAM performs better than DiverTEAM$_\text{post}^-$, indicating that the postprocessing methods can stably improve the solutions.

\section{Conclusion}
This paper addressed the Diversified Top-$k$ Clique search (DT$k$C) and Diversified Top-$k$ Weight Clique search (DT$k$WC) problems, two representatives among various DT$k$ optimization problems, and proposed DiverTEAM, an efficient evolutionary algorithm for solving them. DiverTEAM consists of local search and genetic evolution. The local search focuses on generating high-quality and \textit{diverse} individuals for the population, which will work as a \textit{team} in the genetic algorithm. The genetic algorithm can make the population converge to (near-)optimal solutions efficiently with our designed crossover operator. We further applied a solution-based tabu search technique to prevent the genetic algorithm from searching duplicated solution spaces and proposed some approaches to boost the algorithm performance, including a pseudo graph reduction preprocessing and some postprocessing. 

We compared DiverTEAM with the state-of-the-art heuristic algorithms for DT$k$C and DT$k$WC and conducted experiments on various benchmarks. Extensive experiments show that DiverTEAM exhibits excellent performance and robustness on different benchmarks of both DT$k$C and DT$k$WC. Adequate ablation studies further demonstrate the effectiveness of several key designs in DiverTEAM.


In future work, we will deploy DiverTEAM to other DT$k$ problems to investigate its versatility. Note that the main components of DiverTEAM, including the local search algorithm, the crossover operator, the genetic algorithm, and the postprocessing method, are all performed around the core sub-structure, i.e., clique, of the investigated DT$k$C and DT$k$WC problems. These methods and algorithms could be applied to solve other DT$k$ problems easily by performing them upon the corresponding sub-structures.


\bibliographystyle{named}
\bibliography{ijcai24}

\begin{thebibliography}{}

\bibitem[\protect\citeauthoryear{Albert and Barab{\'a}si}{2002}]{albert2002statistical}
R{\'e}ka Albert and Albert-L{\'a}szl{\'o} Barab{\'a}si.
\newblock Statistical mechanics of complex networks.
\newblock {\em Reviews of modern physics}, 74(1):47, 2002.

\bibitem[\protect\citeauthoryear{Cai and Lin}{2016}]{Weight1}
Shaowei Cai and Jinkun Lin.
\newblock Fast solving maximum weight clique problem in massive graphs.
\newblock In {\em Proceedings of the 25th International Joint Conference on Artificial Intelligence}, pages 568--574, 2016.

\bibitem[\protect\citeauthoryear{Erd{\H{o}}s \bgroup \em et al.\egroup }{1960}]{erdHos1960evolution}
Paul Erd{\H{o}}s, Alfr{\'e}d R{\'e}nyi, et~al.
\newblock On the evolution of random graphs.
\newblock {\em Publication of the Mathematical Institute of the Hungarian Academy of Sciences}, 5(1):17--60, 1960.

\bibitem[\protect\citeauthoryear{Gao \bgroup \em et al.\egroup }{2022}]{XLY22}
Jian Gao, Zhenghang Xu, Ruizhi Li, and Minghao Yin.
\newblock An exact algorithm with new upper bounds for the maximum k-defective clique problem in massive sparse graphs.
\newblock In {\em Proceedings of the 36th {AAAI} Conference on Artificial Intelligence}, pages 10174--10183, 2022.

\bibitem[\protect\citeauthoryear{Huang \bgroup \em et al.\egroup }{2023}]{edge-pattern}
Kai Huang, Haibo Hu, Qingqing Ye, Kai Tian, Bolong Zheng, and Xiaofang Zhou.
\newblock {TED:} towards discovering top-k edge-diversified patterns in a graph database.
\newblock {\em Proceedings of the {ACM} on Management of Data}, 1(1):51:1--51:26, 2023.

\bibitem[\protect\citeauthoryear{Jiang \bgroup \em et al.\egroup }{2018}]{JiangLLM18}
Hua Jiang, Chu{-}Min Li, Yanli Liu, and Felip Many{\`{a}}.
\newblock A two-stage maxsat reasoning approach for the maximum weight clique problem.
\newblock In {\em Proceedings of the 32nd {AAAI} Conference on Artificial Intelligence, the 30th innovative Applications of Artificial Intelligence, and the 8th {AAAI} Symposium on Educational Advances in Artificial Intelligence (EAAI-18)}, pages 1338--1346, 2018.

\bibitem[\protect\citeauthoryear{Luo \bgroup \em et al.\egroup }{2022}]{LLZ+22}
Zihan Luo, Lei Li, Mengxuan Zhang, Wen Hua, Yehong Xu, and Xiaofang Zhou.
\newblock Diversified top-k route planning in road network.
\newblock {\em Proceedings of the VLDB Endowment}, 15(11):3199--3212, 2022.

\bibitem[\protect\citeauthoryear{Lyu \bgroup \em et al.\egroup }{2022}]{LQL+22}
Bingqing Lyu, Lu~Qin, Xuemin Lin, Ying Zhang, Zhengping Qian, and Jingren Zhou.
\newblock Maximum and top-k diversified biclique search at scale.
\newblock {\em The {VLDB} Journal}, 31(6):1365--1389, 2022.

\bibitem[\protect\citeauthoryear{Rossi and Ahmed}{2015}]{RA15}
Ryan~A. Rossi and Nesreen~K. Ahmed.
\newblock The network data repository with interactive graph analytics and visualization.
\newblock In {\em Proceedings of the 29th {AAAI} Conference on Artificial Intelligence}, pages 4292--4293, 2015.

\bibitem[\protect\citeauthoryear{Shahrivari \bgroup \em et al.\egroup }{2020}]{SOP+20}
Hamid Shahrivari, Matthaios Olma, Odysseas Papapetrou, Dimitrios Skoutas, and Anastasia Ailamaki.
\newblock A parallel and distributed approach for diversified top-k best region search.
\newblock In {\em Proceedings of the 23rd International Conference on Extending Database Technology}, pages 265--276, 2020.

\bibitem[\protect\citeauthoryear{Su \bgroup \em et al.\egroup }{2020}]{SuTabu}
Zhouxing Su, Shihao Huang, Chungen Li, and Zhipeng L{\"{u}}.
\newblock A two-stage matheuristic algorithm for classical inventory routing problem.
\newblock In {\em Proceedings of the 29th International Joint Conference on Artificial Intelligence}, pages 3430--3436, 2020.

\bibitem[\protect\citeauthoryear{Sun \bgroup \em et al.\egroup }{2022}]{SWW+22}
Renjie Sun, Yanping Wu, and Xiaoyang Wang.
\newblock Diversified top-r community search in geo-social network: {A} k-truss based model.
\newblock In {\em Proceedings of the 25th International Conference on Extending Database Technology}, pages 2:445--2:448, 2022.

\bibitem[\protect\citeauthoryear{Wang and Zhan}{2018}]{WZ18}
Xin Wang and Huayi Zhan.
\newblock Approximating diversified top-k graph pattern matching.
\newblock In {\em Proceedings of the 29th International Conference on Database and Expert Systems Applications}, volume 11029, pages 407--423, 2018.

\bibitem[\protect\citeauthoryear{Wang \bgroup \em et al.\egroup }{2020}]{Weight2}
Yiyuan Wang, Shaowei Cai, Jiejiang Chen, and Minghao Yin.
\newblock Sccwalk: An efficient local search algorithm and its improvements for maximum weight clique problem.
\newblock {\em Artificial Intelligence}, 280:103230, 2020.

\bibitem[\protect\citeauthoryear{Wu and Yin}{2021}]{wu2021restart}
Jun Wu and Minghao Yin.
\newblock A restart local search for solving diversified top-k weight clique search problem.
\newblock {\em Mathematics}, 9(21):2674, 2021.

\bibitem[\protect\citeauthoryear{Wu \bgroup \em et al.\egroup }{2020}]{wu2020local}
Jun Wu, Chu-Min Li, Lu~Jiang, Junping Zhou, and Minghao Yin.
\newblock Local search for diversified top-k clique search problem.
\newblock {\em Computers \& Operations Research}, 116:104867, 2020.

\bibitem[\protect\citeauthoryear{Wu \bgroup \em et al.\egroup }{2022}]{wu2022head}
Jun Wu, Chu~Min Li, Yupeng Zhou, Minghao Yin, Xin Xu, and Dangdang Niu.
\newblock {HEA-D:} {A} hybrid evolutionary algorithm for diversified top-k weight clique search problem.
\newblock In {\em Proceedings of the 31st International Joint Conference on Artificial Intelligence}, pages 4821--4827, 2022.

\bibitem[\protect\citeauthoryear{Yang \bgroup \em et al.\egroup }{2016}]{YFL16}
Zhengwei Yang, Ada~Wai{-}Chee Fu, and Ruifeng Liu.
\newblock Diversified top-k subgraph querying in a large graph.
\newblock In {\em Proceedings of the 2016 International Conference on Management of Data}, pages 1167--1182, 2016.

\bibitem[\protect\citeauthoryear{Yuan \bgroup \em et al.\egroup }{2016}]{yuan2016diversified}
Long Yuan, Lu~Qin, Xuemin Lin, Lijun Chang, and Wenjie Zhang.
\newblock Diversified top-k clique search.
\newblock {\em The {VLDB} Journal}, 25:171--196, 2016.

\bibitem[\protect\citeauthoryear{Zhang \bgroup \em et al.\egroup }{2022}]{ZSL+22}
Qingyun Zhang, Zhouxing Su, Zhipeng L{\"{u}}, and Lingxiao Yang.
\newblock A weighting-based tabu search algorithm for the p-next center problem.
\newblock In {\em Proceedings of the 31st International Joint Conference on Artificial Intelligence}, pages 4828--4834, 2022.

\bibitem[\protect\citeauthoryear{Zhou \bgroup \em et al.\egroup }{2021a}]{zhou2021solving}
Junping Zhou, Chumin Li, Yupeng Zhou, Mingyang Li, Lili Liang, and Jianan Wang.
\newblock Solving diversified top-k weight clique search problem.
\newblock {\em Science China Information Sciences}, 64(5), 2021.

\bibitem[\protect\citeauthoryear{Zhou \bgroup \em et al.\egroup }{2021b}]{ZHX+21}
Yi~Zhou, Shan Hu, Mingyu Xiao, and Zhang{-}Hua Fu.
\newblock Improving maximum k-plex solver via second-order reduction and graph color bounding.
\newblock In {\em Proceedings of the 35th {AAAI} Conference on Artificial Intelligence, {AAAI} 2021}, pages 12453--12460, 2021.

\bibitem[\protect\citeauthoryear{Zhou \bgroup \em et al.\egroup }{2022}]{Zhou0ZL022}
Jianrong Zhou, Kun He, Jiongzhi Zheng, Chu{-}Min Li, and Yanli Liu.
\newblock A strengthened branch and bound algorithm for the maximum common (connected) subgraph problem.
\newblock In {\em Proceedings of the 31st International Joint Conference on Artificial Intelligence}, pages 1908--1914, 2022.

\end{thebibliography}

\end{document}